\documentclass[11pt]{article}

\usepackage[preprint]{acl}

\usepackage{times}
\usepackage{latexsym}
\usepackage[T1]{fontenc}
\usepackage[utf8]{inputenc}
\usepackage{microtype}
\usepackage{inconsolata}
\usepackage{graphicx}
\usepackage{capt-of}     
\usepackage{float}       
\usepackage{booktabs}    
\usepackage{xcolor}
\usepackage{titletoc}    
\newcommand{\rqanswer}[2]{\par\smallskip\noindent\colorbox{gray!15}{\parbox{\dimexpr\linewidth-2\fboxsep}{\textbf{Answer to RQ#1:} #2}}\par\smallskip}

\title{Termination Lag: Coding Agents Solve Early but Stop Late}

\author{
  \textbf{Musa Cim, Burak Topcu, and Mahmut Taylan Kandemir} \\
  The Pennsylvania State University, University Park, PA, USA \\
  \{mtc5693,bvt5283,mtk2\}@psu.edu
}

\begin{document}
\maketitle

\begin{abstract}
Coding agents can solve a task well before they stop working on it. We characterize \emph{termination lag}—the number of steps between an agent's first correct patch and the end of its run—across five models, two agent frameworks, and three SWE-bench datasets by recording and testing every distinct repository patch generated during execution. Our analysis reveals that 38--82\% of successful runs achieve a correct solution on their initial source-code edit. However,  agents continue to expend substantial effort after correct patches, with 
post-solution steps accounting for a median of 35--37\% for Qwen3-Coder and 54--60\% for Laguna-XS.2 in successful runs on SWE-bench Verified and Multilingual. Agents spend most of these additional steps verifying their work and exploring the repository. In a small fraction of cases (1–8\%), this continued activity leads agents to undo their earlier correct patch, leaving the final solution incorrect.

\end{abstract}

\section{Introduction}
Large language models (LLMs) have enabled autonomous software engineering agents~\citep{react, codeagent, agentless} that address repository-level coding tasks. Given a natural-language issue description and a code repository, these agents inspect source files, execute commands and tests, and modify code to produce a patch that resolves the problem. Their performance is commonly evaluated using benchmarks such as SWE-bench~\citep{verified}, which assess whether the final submitted patch passes task-specific tests.


The primary metric for these agents is the correctness of the final patch, as determined by benchmark scoring. However, this final score obscures when the correct solution was actually obtained, which is a critical driver of execution duration and overall cost. Consider, for instance, two agents that apply the same correct edit at step 10. While one agent submits the patch immediately, the other executes 40 additional steps to rerun tests, inspect callers, and review its diff before halting. Both receive identical credit, even though the early-solving agent is more efficient than the other. Furthermore, an agent may produce a correct patch midway through a run and later even overwrite and corrupt it, so a failed submission can hide a task that the agent has already solved~\citep{Coherence}. Final-patch evaluation therefore conflates two distinct abilities: producing a correct fix and recognizing that the fix is complete.

We argue that knowing precisely when to terminate execution is just as critical to agent performance as generating the correct fix. Every additional agent step leads to another model call over a growing context, adding tokens, latency, and compute, in addition to introducing the risk of breaking code that already works. Yet some verification is necessary, since the agent cannot see the benchmark tests and must rely on its own evidence to judge whether the patch is correct. Characterizing stopping behavior thus requires knowing when a correct patch first exists, how many post-solution steps follow, and what these steps actually do to the patch.

To address these questions, we conduct an empirical analysis of agent trajectories, tracking each run from the first source-code edit, through the generation of the correct patch, to final termination. We refer to the number of steps until the patch first passes benchmark tests as \emph{time to first correct patch} (TTFCP), and the steps the agent takes after that point as \emph{termination lag} (TL). We record the repository patch after every agent step and evaluate each distinct patch offline against the official benchmark tests, which the agent never sees, so the measurement does not influence its behavior. Our core evaluation covers Qwen3-Coder and Laguna-XS.2 on SWE-bench Verified and Multilingual; we further test generalization with Claude Haiku 5.5 and GPT-6 Luna on Verified, with Seed-OSS on SWE-bench Pro, and additionally with the Mini-SWE agent framework.

Correct patches appear early: on SWE-bench Verified, Qwen3-Coder's first source-code edit already passes the benchmark tests in 82\% of runs that reach a correct patch, and the rate is 38--82\% across all settings. Agents then keep working well past this point: in successful runs, a median of 35--60\% of steps come after the first correct patch in our core settings. This termination lag also appears for Claude Haiku 5.5 and GPT-6 Luna, for Seed-OSS on SWE-bench Pro, and under a second agent framework. Agents spend most of these steps checking their work: verification accounts for 48--63\% and exploration for another 15--29\%, while further source-code edits are rare in most settings. Most runs keep the correct patch to the end (61--96\%), but in 1--8\% later edits break it and the agent submits an incorrect patch, a failure that stopping at the first correct patch would have avoided.

These results show that the final solve rate hides both how early agents solve a task and how much work follows. A practical stopping rule would need to decide when to submit using only the evidence available during the run. Our measurements give a reference point for such rules: termination lag bounds how much work stopping at the right moment could save, and the runs that break a correct fix show what continuing can cost.

\section{Related Work}

\subsection{Overthinking and Excess Generation}
Reasoning models often reach the right answer early and keep writing~\citep{overthinking}, and code models keep
generating after the target function is complete; both can be cut short, by a learned stopping
signal~\citep{excessgeneration} or by stopping once the output passes the tests~\citep{greenai,Babbling}. These approaches focus on reducing unnecessary generation within a single model response.

\subsection{Cost and Inefficiency of Coding Agents}
Several studies characterize where agent effort goes: token cost varies with programming language~\citep{tokenmaxing},
prompts that ask for certainty add extra verification~\citep{promptinduced}, failed runs are longer and keep going after
they become unrecoverable~\citep{failtrajec,failureasproces}, some passing runs keep exploring after a verified
fix~\citep{agentlens}, and agents' own tests often accept wrong patches~\citep{mohsin}. However, these studies do not systematically quantify how many steps coding agents take between their first correct repository-level patch and termination.

\subsection{Early Termination, Budgets and Restart}
Many works stop agent runs early, mostly by predicting failure~\citep{agentstop,failfast} or
success~\citep{eet,EarlyEval} from the trajectory, or by capping turns and cost~\citep{moreless,TokenEconomizer}; recent
studies question how well such predictions support stopping or restarting decisions~\citep{contrestart,monitor}. These
methods predict success or failure; we measure when the patch is already correct, giving the most that stopping on
success can save.

\subsection{Intermediate Patches}
Several concurrent studies examine intermediate patches in coding-agent runs. \citet{Coherence} show that a correct fix written mid-run can be lost to later edits. \citet{monitor} test the patch after each action on 19 tasks to guide restart decisions, while \citet{execprefix} resume runs from intermediate states to estimate the probability of eventual success from each state. These studies examine different aspects of how correctness evolves throughout coding-agent execution.

\section{Methodology}

\subsection{Agent Framework}
We use OpenHands~\citep{openhands} as the primary agent framework for all five models. It provides tools for inspecting repository
files, editing code, and executing commands and tests. To examine whether our findings extend to another framework,
we also evaluate Qwen3-Coder on SWE-bench Verified using mini-SWE-agent~\citep{yang2024sweagent} (Appendix~\ref{sec:minisweagent}).

\subsection{Closed-Source Models}
We evaluate two closed-source models, Claude Haiku 5.5 and GPT-6 Luna, accessed through the Anthropic and OpenAI
APIs, respectively. Both models are evaluated on SWE-bench Verified using OpenHands. Each run continues until the agent
submits its patch, with a budget of 100 steps.

\subsection{Open-Source Models}
We evaluate three open-source models served through vLLM on Aurora~\citep{alcf_aurora}: Qwen3-Coder-30B-A3B, Laguna-XS.2, and
Seed-OSS-36B. Qwen3-Coder and Laguna-XS.2 are evaluated on the complete SWE-bench Verified and Multilingual
benchmarks, with budgets of 100 and 250 steps and context windows of 131K and 256K tokens, respectively. To assess
whether our findings extend to an additional model and more challenging tasks, we evaluate Seed-OSS-36B on SWE-bench
Pro with a budget of 150 steps and a 131K-token context window.
We use Qwen3-Coder and Laguna-XS.2 on SWE-bench Verified and Multilingual as our core evaluation throughout the paper.

\subsection{Benchmarks}
We select three benchmarks to assess whether our findings generalize from Python tasks to other programming
languages and to more challenging, longer-horizon tasks. SWE-bench Verified~\citep{verified} provides 500 tasks from Python
repositories, while SWE-bench Multilingual~\citep{multilingual} extends the evaluation to 300 tasks spanning nine programming languages.
SWE-bench Pro~\citep{swepro} further broadens the evaluation to harder tasks requiring more sustained agent execution.

\subsection{Recording Patches After Each Agent Step}
To determine when a correct patch first appears, we save the current patch after each agent step, encompassing the
model's reasoning, its selected action, and any resulting tool execution. A logging proxy between the agent and the
model endpoint records the patch before forwarding the next model request, while the agent is waiting for a
response. Each recorded patch contains the difference between the current repository state and the task's base
commit. We construct each patch using a temporary Git index, leaving the agent's working repository unchanged. After
execution, we evaluate the recorded patches in isolated containers using the benchmark's official evaluation
procedure, evaluating identical patches only once. The agent receives no feedback from this retrospective
evaluation, allowing us to track correctness without influencing its subsequent decisions.

\subsection{Time to First Correct Patch and Termination Lag}
We use two metrics to characterize when a correct patch appears relative to when execution ends: Time to First
Correct Patch (TTFCP) and Termination Lag (TL). For a trajectory containing $S$ agent steps, let $F$ denote the
earliest step whose patch passes the official benchmark evaluation; we define TTFCP, measured in agent steps, as
\begin{equation}
\mathrm{TTFCP} = F. \label{eq:ttfcp}
\end{equation}
We define termination lag, the number of additional steps executed after the first correct patch appears, as
\begin{equation}
\mathrm{TL} = S - F. \label{eq:tl}
\end{equation}
To compare trajectories of different lengths, we define Normalized Termination Lag, the fraction of execution
occurring after the first correct patch appears, as
\begin{equation}
\mathrm{NTL} = \frac{S - F}{S}. \label{eq:ntl}
\end{equation}
Here, $S$ is the step at which the agent stops on its own. Table~\ref{tab:metrics} reports these metrics for
trajectories that end with a correct patch. We also define the first patch as the patch recorded after the first
source-code edit and count the distinct patches up to and including the first correct one, over all trajectories
that reach a correct patch.

\begin{figure*}[t]  
\centering
\includegraphics[width=0.7\textwidth]{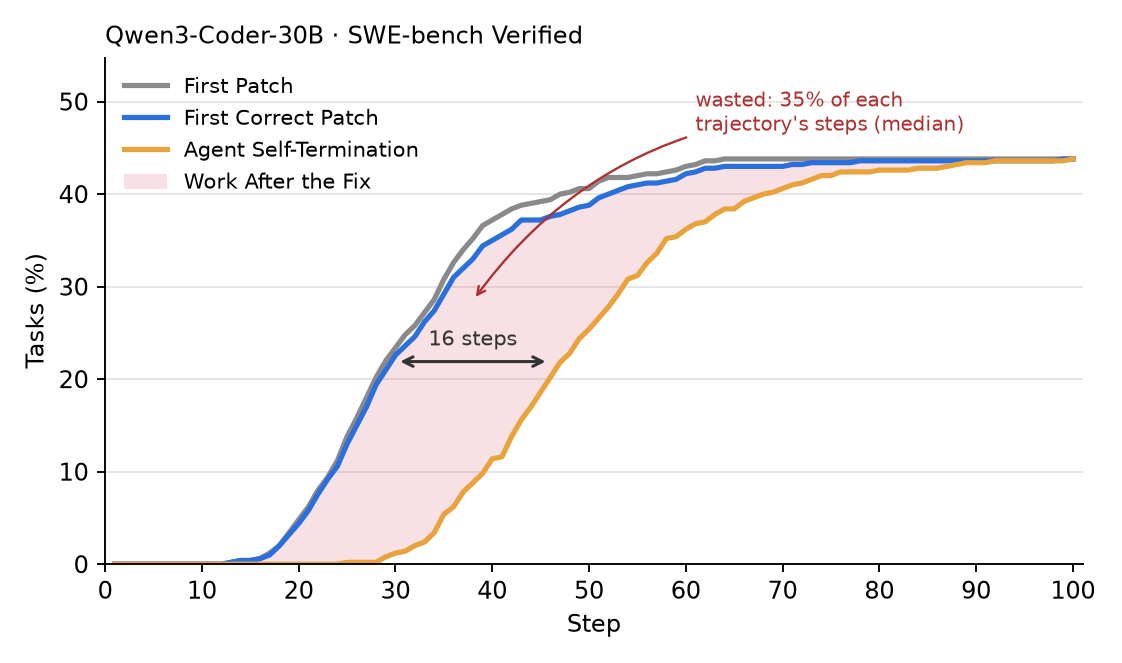}
\vspace{-3mm}
\caption{First correct patch and agent stop by step, Qwen3-Coder-30B, SWE-bench Verified.}
\label{fig:waste}
\vspace{-4mm}
\end{figure*}

\section{Evaluation}
\label{sec:waste}

\subsection{Research Questions}
\label{sec:rqs}
Our analysis addresses the following research questions.

\noindent\textbf{RQ1 (First edit):} When does the agent make its first source-code edit, and how often is that edit
already correct?

\noindent\textbf{RQ2 (Termination lag):} How long does the agent keep working after its first correct patch?

\noindent\textbf{RQ3 (Post-fix activity):} What does the agent do in these additional steps?

\noindent\textbf{RQ4 (Effect on correctness):} Does the additional work help or hurt the patch?

\subsection{Correct Patches Appear Early}
\label{sec:tl}

Coding agents approach software tasks through repeated cycles of code inspection, editing, and testing before
deciding when to stop. During this process, a correct patch may emerge before the agent considers its work complete.
We investigate the relationship between these two milestones by examining when correct patches first appear and how
much execution continues before termination. For solved trajectories, we compare the step at
which correctness is first achieved with the final step of execution. Figure~\ref{fig:waste} presents this
comparison for Qwen3-Coder on SWE-bench Verified. At each step, the blue curve shows the percentage of tasks for
which a correct patch has already appeared, while the orange curve shows the percentage for which execution has
already ended with a correct patch. The blue curve rises earlier than the orange curve, indicating
that correct patches often appear well before execution ends. Among solved tasks, the median
termination lag is 16 steps, and the median normalized termination lag is 35\%. Thus, a substantial portion of these
trajectories occurs after the agent has already produced a correct patch.

We next examine whether this separation appears with other models and benchmarks. On SWE-bench Multilingual,
Qwen3-Coder exhibits a median termination lag of 18 steps and a median normalized termination lag of 37\%
(Appendix~\ref{sec:waste_all}, Figure~\ref{fig:waste_qwen_ml}). For Laguna-XS.2, these values reach 57 steps and
60\% on Verified (Appendix~\ref{sec:waste_all}, Figure~\ref{fig:waste_laguna_v}), and 56 steps and 54\% on
Multilingual (Appendix~\ref{sec:waste_all}, Figure~\ref{fig:waste_laguna_ml}). The separation also appears in the
commercial models. For Claude Haiku 5.5 on Verified, the median first-correct-patch step is 9, while the
median termination lag is 10 steps and the median normalized termination lag is 54\% (Appendix~\ref{sec:waste_all},
Figure~\ref{fig:waste_haiku}). GPT-6 Luna exhibits a median termination lag of 6 steps and a median normalized
termination lag of 36\% on the same benchmark (Appendix~\ref{sec:waste_all}, Figure~\ref{fig:waste_gpt6luna}). The
same pattern appears for Seed-OSS-36B on SWE-bench Pro (Appendix~\ref{sec:waste_all},
Figure~\ref{fig:waste_seed_pro}). To test another agent framework, we repeat the
Qwen3-Coder evaluation on Verified using mini-SWE-agent. This setting yields a median termination lag of 12 steps
and a median normalized termination lag of 37\%
(Appendix~\ref{sec:waste_all}, Figure~\ref{fig:waste_msa} and Table~\ref{tab:msa}). To quantify the potential
benefit of earlier termination, we consider an oracle that stops each run at its first correct patch, leaving runs
that never reach correctness unchanged. Among solved runs, this would eliminate a median of 35\% and
37\% of each trajectory's steps for Qwen3-Coder on Verified and Multilingual, respectively, 60\% and 54\% for
Laguna-XS.2, and 54\% for Haiku on Verified, while preserving every originally solved task. These results highlight
the potential of closing the gap between producing a correct patch and terminating.
\rqanswer{2}{Across three benchmarks, five models, and two agent frameworks, agents consistently keep working well
after their first correct patch: in successful runs, roughly a third to more than half of the steps come after it.}

\begin{table}[h]
\centering
\caption{Termination lag in the core settings. All rows below Solved are medians over solved runs; TTFCP, TL and
NTL follow Eqs.~\ref{eq:ttfcp}--\ref{eq:ntl}. Context growth: relative increase in the model's input context, in
tokens, from the first correct patch to the final step, $(\mathrm{ctx}_S-\mathrm{ctx}_F)/\mathrm{ctx}_F$.}
\label{tab:metrics}
\small
\setlength{\tabcolsep}{0pt}
\begin{tabular*}{\columnwidth}{@{\extracolsep{\fill}}lcccc@{}}
\toprule
 & \multicolumn{2}{c}{\textbf{Qwen3-Coder}} & \multicolumn{2}{c}{\textbf{Laguna-XS.2}} \\
\cmidrule(lr){2-3}\cmidrule(lr){4-5}
 & Verif. & Multi. & Verif. & Multi. \\
\midrule
Solved (\%)        & 43.8 & 22.3 & 56.0 & 48.0 \\
TTFCP              & 30   & 34   & 38   & 50   \\
Final step $S$     & 48   & 54   & 95   & 108  \\
TL                 & 16   & 18   & 57   & 56   \\
\midrule
NTL (\%)           & 35   & 37   & 60   & 54   \\
Context growth (\%) & \textcolor{red}{+30}  & \textcolor{red}{+28}  & \textcolor{red}{+78}  & \textcolor{red}{+55}  \\ 
\bottomrule
\end{tabular*}
\end{table}

\twocolumn[{ 
\centering
\captionof{table}{From the first patch to the first correct patch, in runs that reach a correct patch. First patch:
the first step that edits the repository's source. Attempts: distinct patches up to and including the first correct
one.}
\label{tab:firstpatch}
\begin{tabular*}{\textwidth}{@{\extracolsep{\fill}}lcccccc}
\toprule
\emph{Model} & \multicolumn{2}{c}{\textbf{Qwen3-C.}} & \multicolumn{2}{c}{\textbf{Laguna-XS.2}} & \textbf{GPT-6 Luna} & \textbf{Seed-OSS} \\
\cmidrule(lr){2-3}\cmidrule(lr){4-5}\cmidrule(lr){6-6}\cmidrule(lr){7-7}
\emph{Benchmark} & Verified & Multi. & Verified & Multi. & Verified & Pro \\
\midrule
First patch (median step)                  & 30 & 34 & 36 & 45 & 9 & 17 \\
First correct patch (median step)          & 30 & 34 & 38 & 51 & 9 & 26 \\
First patch already correct                & 82\% & 66\% & 70\% & 48\% & 80\% & 38\% \\
Attempts until correct (mean) & 1.3 & 1.7 & 1.9 & 2.8 & 1.7 & 3.6 \\
\bottomrule
\end{tabular*}
\medskip
\captionof{table}{What agents do after their first correct patch. Top: share of the steps after the fix, by kind.
Bottom: what happens to the patch in runs that reach a correct patch.}
\label{tab:post}
\begin{tabular*}{\textwidth}{@{\extracolsep{\fill}}lcccccc}
\toprule
\emph{Model} & \multicolumn{2}{c}{\textbf{Qwen3-Coder}} & \multicolumn{2}{c}{\textbf{Laguna-XS.2}} & \textbf{GPT-6 Luna} & \textbf{Seed-OSS} \\
\cmidrule(lr){2-3}\cmidrule(lr){4-5}\cmidrule(lr){6-6}\cmidrule(lr){7-7}
\emph{Benchmark} & Verified & Multi. & Verified & Multi. & Verified & Pro \\
\midrule
\multicolumn{7}{l}{\emph{Steps after the fix}} \\
\quad Verify                    & 63\% & 62\% & 59\% & 48\% & 55\% & 50\% \\
\quad Explore                   & 21\% & 15\% & 29\% & 29\% & 22\% & 22\% \\
\quad Edit the source again     & 2\% & 5\% & 4\% & 6\% & 5\% & 19\% \\
\quad Wrap-up                   & 6\% & 7\% & 6\% & 9\% & 12\% & 7\% \\
\quad Think, other              & 7\% & 12\% & 3\% & 8\% & 6\% & 3\% \\
\midrule
\multicolumn{7}{l}{\emph{The patch after the fix (share of runs)}} \\
\quad Stays correct to the end  & 96\% & 82\% & 84\% & 61\% & 88\% & 79\% \\
\quad Breaks, correct again     & 3\% & 10\% & 11\% & 31\% & 8\% & 17\% \\
\quad Breaks, run ends unsolved & 1\% & 8\% & 5\% & 8\% & 4\% & 4\% \\
\bottomrule
\end{tabular*}
\medskip
}]
 
\section{Understanding Agent Behavior During Termination Lag}
\label{sec:why}

In the previous section, we established the presence of termination lag and quantified its extent across the
evaluated settings. We now examine what happens during this interval to better understand the behavior underlying
these additional steps. Specifically, we investigate how agents spend their steps after producing a correct patch and
how their subsequent actions affect the correctness of that patch. For trajectories that end successfully, we
classify each step after the first correct patch according to the activity performed, using the recorded tools and
their arguments. We group these steps into verification, exploration, source-code editing, wrap-up, and reasoning or
other activities (Table~\ref{tab:post}; Appendix Figure~\ref{fig:post_actions}). Separately, we track patch
correctness in every trajectory that reaches a correct patch, including those that subsequently end unsuccessfully.
Verification includes running existing repository tests, writing and executing additional test scripts, and building
the project. Exploration includes inspecting files, searching the repository, and examining changes through Git
commands, while source-code editing covers further modifications to the implementation. Wrap-up includes removing
temporary files and submitting the final response; reasoning and other activities cover the remaining steps.

\paragraph{From the First Patch to the First Correct Patch.}
To better understand when correct solutions emerge, we examine whether agents produce a correct patch on their first
attempt or require further revisions. We define the first patch as the patch recorded after the first source-code
edit and conduct this analysis over all trajectories that reach a correct patch. Table~\ref{tab:firstpatch} reports
when the first and first correct patches appear, the proportion of first patches that are already correct, and the
mean number of distinct patches produced up to and including the first correct one.
Among trajectories that reach a correct patch, Qwen3-Coder's first patch is already correct in 82\% of runs on
Verified and 66\% on Multilingual. The corresponding proportions are 70\% and 48\% for Laguna-XS.2, 80\% for GPT-6
Luna on Verified, and 38\% for Seed-OSS on Pro. Across these settings, agents produce an average of 1.3--3.6 distinct
patches before reaching correctness, including the first correct patch itself.
For Qwen3-Coder, the median first-patch and first-correct-patch steps coincide at 30 on Verified and 34 on
Multilingual; for GPT-6 Luna on Verified, both medians are 9. For Laguna-XS.2, the corresponding medians are 36 and
38 on Verified and 45 and 51 on Multilingual, while Seed-OSS on Pro records medians of 17 and 26. Together with the
first-patch correctness rates, these results show that many trajectories reach correctness with their initial
source-code edit, although further revisions are needed in a substantial proportion of runs.

\rqanswer{1}{Agents make their first source-code edit at a median step of 9--45. In runs that reach a correct patch,
this first edit is already correct in 38--82\% of cases, and agents need 1.3--3.6 distinct patches on average to reach
correctness.}

\paragraph{Where the Steps Go.}
Verification is the largest activity category in every evaluated setting, accounting for 48--63\% of steps during
termination lag (Table~\ref{tab:post}). These results show that agents devote much of their continued
execution to additional verification after a benchmark-correct patch has already appeared. Exploration accounts for
another 15--29\% of steps during termination lag, as agents continue inspecting files, searching code, and reviewing
repository changes. Source-code editing is less frequent for Qwen3-Coder, Laguna-XS.2, and GPT-6 Luna, representing
2--6\% of steps across their evaluated settings. Seed-OSS on SWE-bench Pro has a source-code editing share of 19\%
(Table~\ref{tab:post}).

\rqanswer{3}{Verification (running tests, writing checks, building) takes 48--63\% of the steps after the first
correct patch, and exploration another 15--29\%. Further source edits are rare (2--6\%), except for Seed-OSS on SWE-bench
Pro (19\%).}

\paragraph{What Happens to the Correct Patch.}
We next examine how patch correctness evolves after the first correct patch appears. We distinguish three outcomes
after a correct patch first appears: it remains correct through the end, later becomes incorrect but is repaired
before termination, or later becomes incorrect and the run ends without recovering the earlier correct solution.
Across the evaluated settings, 61--96\% of trajectories that reach a correct patch maintain correctness throughout
the remaining execution (Table~\ref{tab:post}). A correct patch becomes incorrect and is subsequently repaired in
11\% of Laguna-XS.2 trajectories on Verified, 31\% on Multilingual, and 17\% of Seed-OSS trajectories on Pro
(Table~\ref{tab:post}). Across the evaluated settings, 1--8\% of trajectories that reach a correct patch ultimately
end with an incorrect patch. In these cases, stopping at the first correct patch would have preserved benchmark
success.

\rqanswer{4}{Once a patch passes the benchmark tests, further work cannot improve its benchmark correctness but may
introduce regressions: 61--96\% of runs that
reach a correct patch keep it to the end, while 1--8\% break it and end with an incorrect patch that stopping at the
first correct patch would have avoided.}

\paragraph{Putting It Together.}
Verification and exploration account for most steps during termination lag across the evaluated settings.
Although patches often remain correct throughout this interval, further changes can introduce regressions that
require repair or leave the task unsolved at termination. Continued execution therefore incurs additional steps and
can also compromise a solution that already satisfies the benchmark's correctness criteria.

\section{Conclusion}

We measured when coding agents first produce a correct patch and how long they continue afterward. Across three
benchmarks, five models and two agent frameworks, a consistent gap separates solving a task from stopping: in our core
settings, a median of 35--60\% of each solved run's steps come after the first correct patch, spent mostly on
verification and exploration, and later edits sometimes break the fix. Stopping behavior is therefore a distinct part
of agent performance, to be evaluated by when correctness is reached and whether it is kept until the end of the run.

\section*{Limitations}
Our evaluation covers three benchmarks, five models, and two agent frameworks, with each task
run once, and may not transfer to other settings. The few solved runs that reach the step or context limit (1--4\% per
setting) are counted at their final step, and we measure execution in steps, which approximates but does not directly measure computational cost.
Step categories are assigned by rules over the recorded tools and their arguments, and Seed-OSS on SWE-bench Pro
solves few tasks, so its statistics rest on a small sample.

\section*{Ethical Considerations}
The implementation of this work is conducted with transparency, providing full disclosure of all technical details,
limitations, and potential issues to the relevant stakeholders.

\bibliography{custom}

\onecolumn    
\appendix
\startcontents[appendix]
\section*{Appendix Contents}
\printcontents[appendix]{}{1}{\setcounter{tocdepth}{3}}
\clearpage

\section{Generalization}
\label{sec:waste_all}

We repeat the analysis of Section~\ref{sec:waste} for the other models and benchmarks, and with a second agent
framework. Each plot reads as Figure~\ref{fig:waste}.

\subsection{Closed-Source Models}

\subsubsection{Claude Haiku 5.5, SWE-bench Verified}
\label{sec:haiku}
\begin{figure}[H]
\centering
\includegraphics[width=0.8\textwidth]{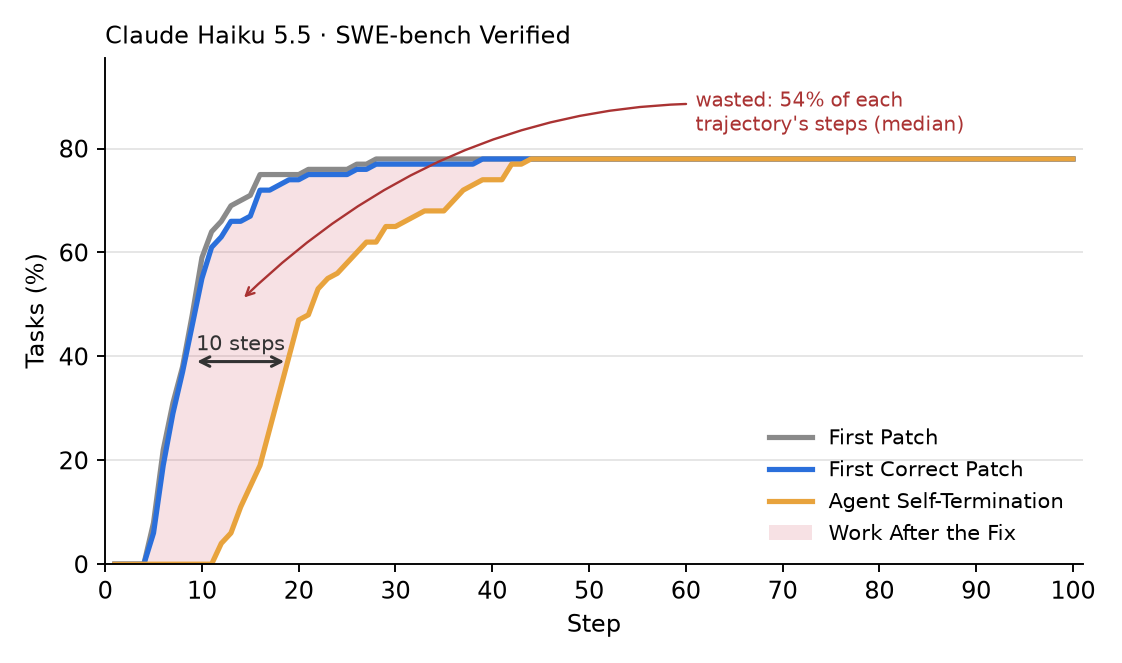}
\caption{First correct patch and agent stop by step, Claude Haiku 5.5, SWE-bench Verified.}
\label{fig:waste_haiku}
\end{figure}

\subsubsection{GPT-6 Luna, SWE-bench Verified}
\begin{figure}[H]
\centering
\includegraphics[width=0.8\textwidth]{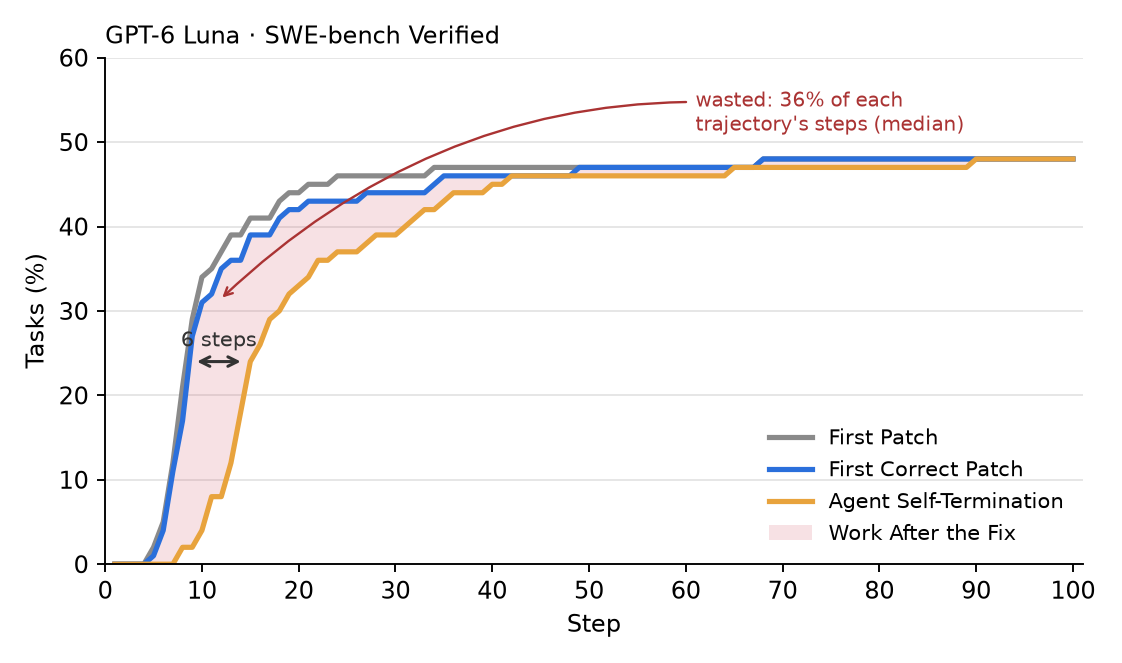}
\caption{First correct patch and agent stop by step, GPT-6 Luna, SWE-bench Verified.}
\label{fig:waste_gpt6luna}
\end{figure}

\subsection{Open-Source Models}

\subsubsection{Qwen3-Coder-30B, SWE-bench Multilingual}
\begin{figure}[H]
\centering
\includegraphics[width=0.8\textwidth]{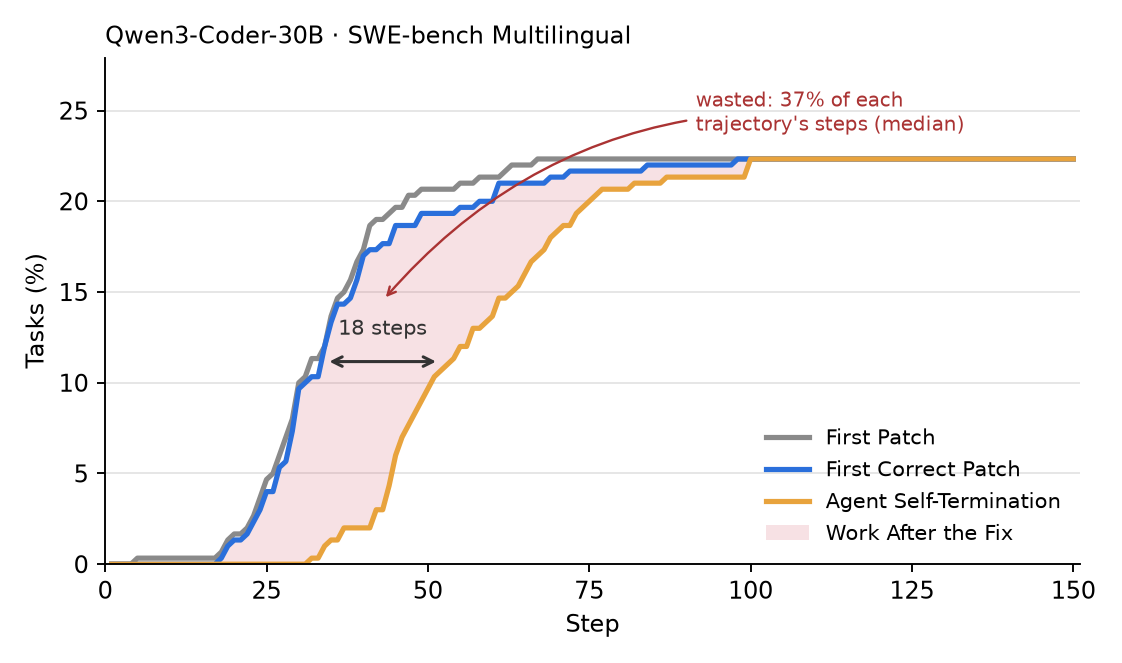}
\caption{First correct patch and agent stop by step, Qwen3-Coder-30B, SWE-bench Multilingual.}
\label{fig:waste_qwen_ml}
\end{figure}

\subsubsection{Laguna-XS.2, SWE-bench Verified}
\begin{figure}[H]
\centering
\includegraphics[width=0.8\textwidth]{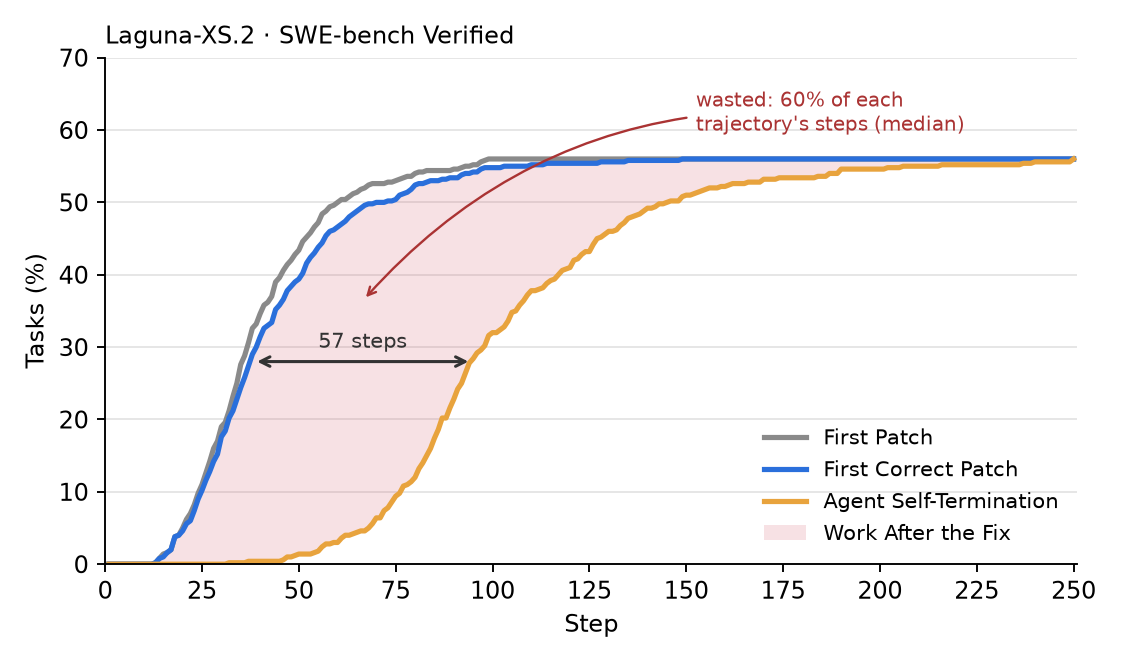}
\caption{First correct patch and agent stop by step, Laguna-XS.2, SWE-bench Verified.}
\label{fig:waste_laguna_v}
\end{figure}

\subsubsection{Laguna-XS.2, SWE-bench Multilingual}
\begin{figure}[H]
\centering
\includegraphics[width=0.8\textwidth]{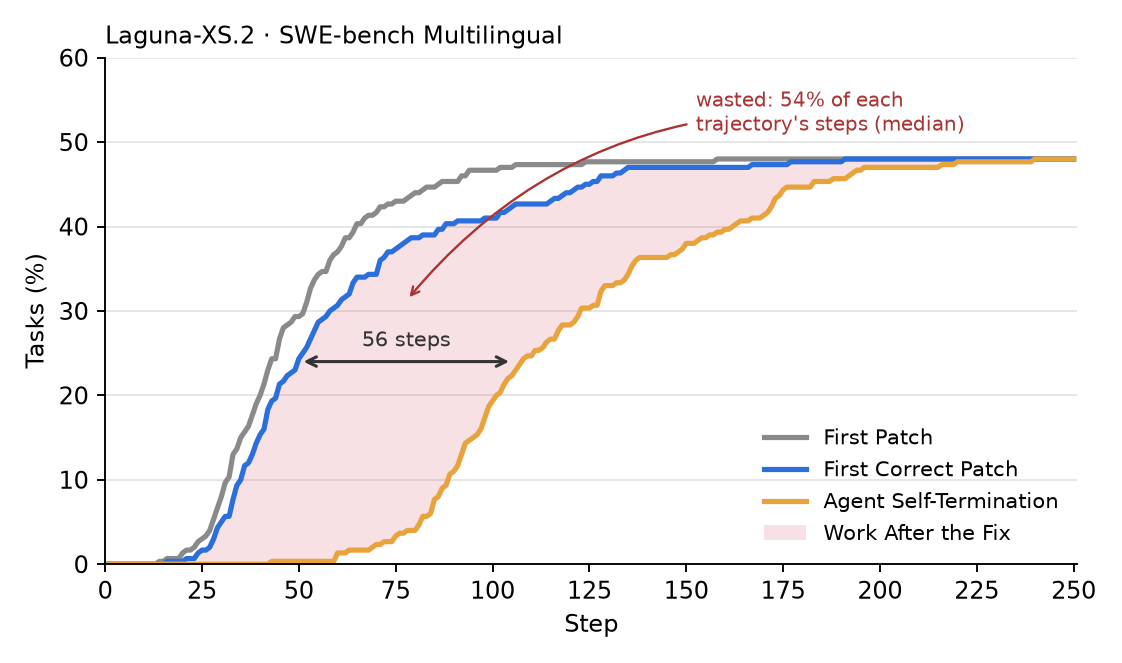}
\caption{First correct patch and agent stop by step, Laguna-XS.2, SWE-bench Multilingual.}
\label{fig:waste_laguna_ml}
\end{figure}

\subsubsection{Seed-OSS-36B, SWE-bench Pro}
\label{sec:seed_pro}

\begin{figure}[H]
\centering
\includegraphics[width=0.8\textwidth]{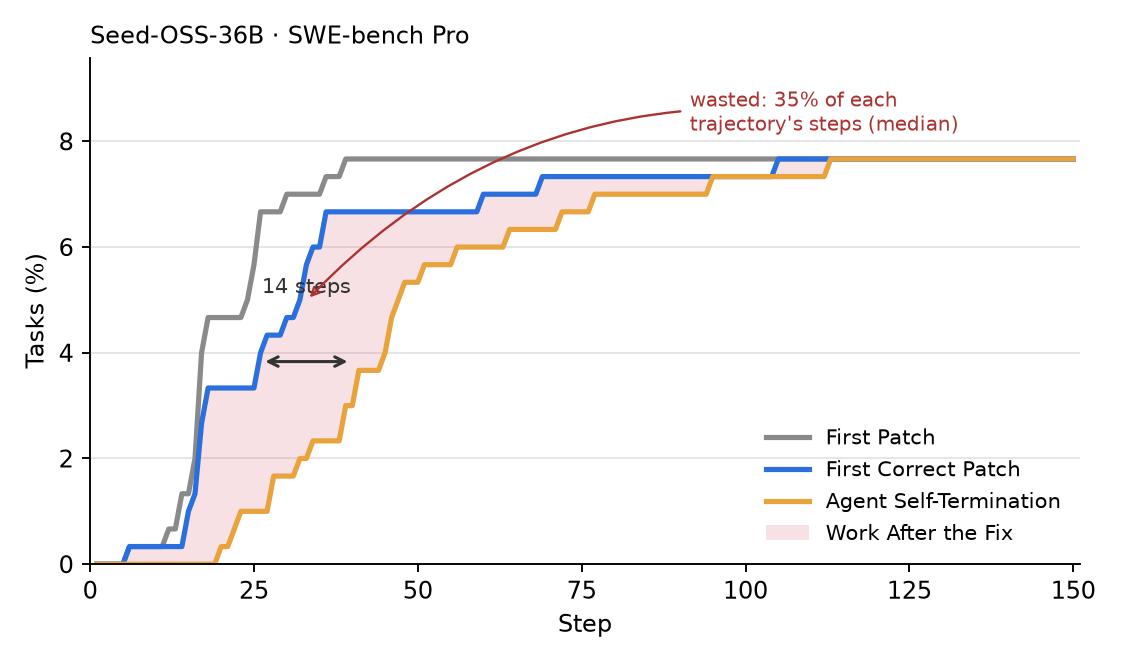}
\caption{First correct patch and agent stop by step, Seed-OSS-36B, SWE-bench Pro.}
\label{fig:waste_seed_pro}
\end{figure}

\subsection{Additional Agent Framework}

\subsubsection{mini-SWE-agent: Qwen3-Coder-30B, SWE-bench Verified}
\label{sec:minisweagent}

We repeat the Qwen3-Coder Verified run with a second agent framework, mini-SWE-agent (bash only, its own SWE-bench
prompt).

\begin{figure}[H]
\centering
\includegraphics[width=0.8\textwidth]{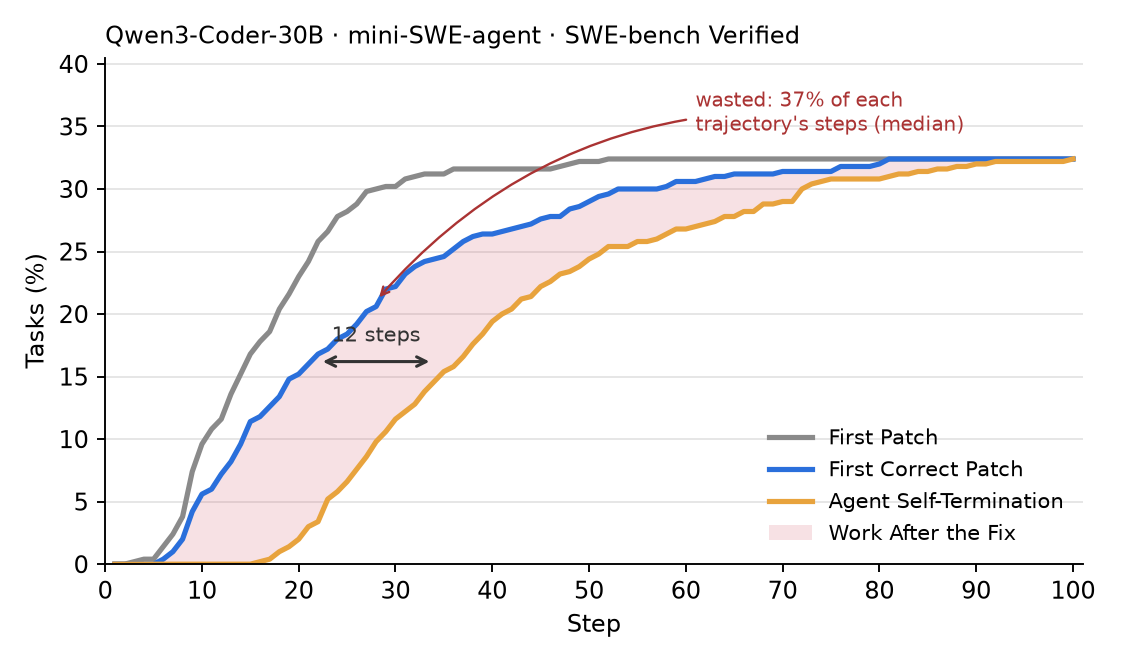}
\caption{First correct patch and agent stop by step, mini-SWE-agent, Qwen3-Coder, Verified.}
\label{fig:waste_msa}
\end{figure}

\begin{table}[H]
\centering
\caption{OpenHands vs.\ mini-SWE-agent, Qwen3-Coder, Verified (500 tasks).}
\label{tab:msa}
\begin{tabular}{|l|r|r|}
\hline
 & \textbf{OpenHands} & \textbf{mini-SWE-agent} \\
\hline
Solved                                            & 43.8\% & 32.4\% \\
First correct step, median (solved runs)          & 30 & 22 \\
Agent's own stop, median (solved runs)            & 48 & 37 \\
Steps after the fix, median / p90                 & 16 / 22 & 12 / 23 \\
Share of the trajectory after the fix, median     & 35\% & 37\% \\
\hline
Steps per task, median                            & 53 & 52 \\
Uncached compute tokens per task, median          & 39.0K & 22.2K \\
Context at the last step, median (tokens)         & 42.9K & 20.5K \\
\hline
\end{tabular}
\end{table}

\section{Actions After the Fix}
\label{sec:post_actions}

\begin{figure}[H]
\centering
\includegraphics[width=0.92\textwidth]{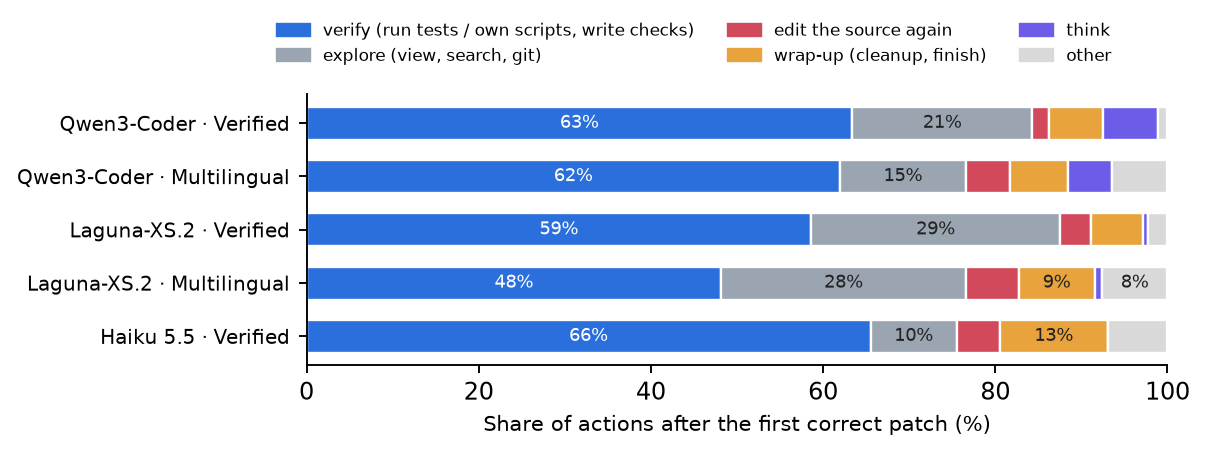}
\caption{Share of actions after the first correct patch, by kind (solved runs).}
\label{fig:post_actions}
\end{figure}

\end{document}